\documentclass[sn-mathphys,Numbered]{sn-jnl}% Math and Physical Sciences Reference Style
\usepackage{graphicx}%
\usepackage{multirow}%
\usepackage{amsmath,amssymb,amsfonts}%
\usepackage{amsthm}%
\usepackage{mathrsfs}%
\usepackage[title]{appendix}%
\usepackage{pifont}% http://ctan.org/pkg/pifont
\usepackage{xcolor}%
\usepackage{textcomp}%
\usepackage{manyfoot}%
\usepackage{booktabs}%
\usepackage{algorithm}%
\usepackage{algorithmicx}%
\usepackage{algpseudocode}%
\usepackage{listings}%
\theoremstyle{thmstyleone}%
\theoremstyle{thmstyletwo}%

\theoremstyle{thmstylethree}%

\begin{document}

\title[Article Title]{ExioML: Eco-economic dataset for Machine Learning in Global Sectoral Sustainability}


\author[1]{\fnm{Yanming} \sur{Guo}}\email{yguo0337@uni.sydney.edu.au}

\author[2]{\fnm{Charles}
\sur{Guan}}\email{charles.guan@richdataco.com}

\author[1]{\fnm{Jin} \sur{Ma}}\email{j.ma@sydney.edu.au}

\affil[1]{\orgdiv{School of Electrical \& Computer Engineering}, \orgname{University of Sydney}}

\affil[2]{\orgname{Rich Data Co}}


\abstract{The Environmental Extended Multi-Regional Input-Output analysis is the predominant framework in Ecological Economics for assessing the environmental impact of economic activities. This paper introduces ExioML, the first Machine Learning benchmark dataset designed for sustainability analysis, aimed at lowering barriers and fostering collaboration between Machine Learning and Ecological Economics research. A crucial greenhouse gas emission regression task was conducted to evaluate sectoral sustainability and demonstrate the usability of the dataset. We compared the performance of traditional shallow models with deep learning models, utilizing a diverse Factor Accounting table and incorporating various categorical and numerical features. Our findings reveal that ExioML, with its high usability, enables deep and ensemble models to achieve low mean square errors, establishing a baseline for future Machine Learning research. Through ExioML, we aim to build a foundational dataset supporting various Machine Learning applications and promote climate actions and sustainable investment decisions.}




\maketitle

\section{Background \& Summary}\label{sec1}
The increase in Greenhouse Gas (GHG) emissions due to fossil-fuel-driven economic development has precipitated a global warming crisis. This concern led to establishing the Paris Agreement in 2015, aiming to limit long-term temperature rise to no more than 2 \(^\circ \)C above pre-industrial levels \cite{schleussner2016science}. Concurrently, the Sustainable Development Goals (SDGs), specifically Goal 13 and Goal 8, were proposed to emphasize taking climate action to reduce emissions while maintaining economic growth. To address such climate-trade dilemma, researchers from various disciplines strive to balance climate action with economic growth and human well-being \cite{rao2014climate, jorgenson2014economic}. Recently, the Machine Learning (ML) technique \cite{lecun2015deep} has emerged as a significant tool for accurate prediction to assist climate change decision-making \cite{rolnick2022tackling}. Specifically, ML algorithms have been explored in aiding nearly real-time global weather forecasting \cite{lam2023learning}, land monitoring via satellite imagery \cite{stanimirova2023global, he2016deep}, and the prediction of disturbances in electric grids \cite{zheng2022multi}. 

\begin{figure} [h!]
    \centering
    \includegraphics[width=0.7\linewidth]{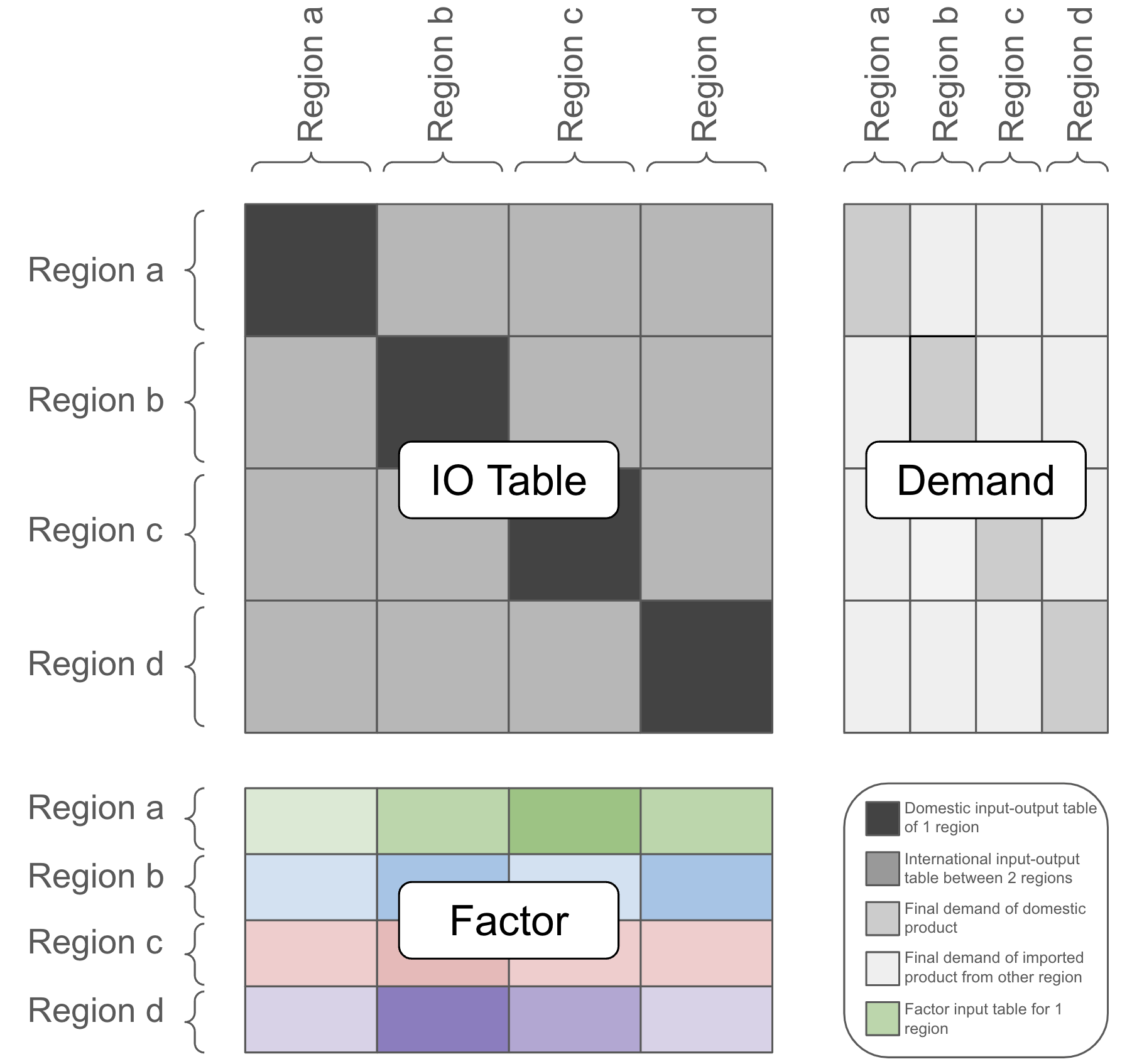}
    \caption{Environmental Extended Multi-Regional Input-Output (EE-MRIO) data are represented in a high-dimensional matrix format, tracking the monetary resource transfers between international sectors using input-output and demand matrices. Additionally, it accounts for the environmental impacts of economic activities through the Factor Accounting table.}
    
    \label{fig:EEMRIO}
\end{figure}

The predominant Ecological Economics (EE) research framework, the Environmentally Extended Multiregional Input-Output (EE-MRIO) analysis, effectively models global economic interactions of sectors within a network structure \cite{leontief1963multiregional, wang2019carbon, sun2020exploring, steinberger2012pathways, jakob2013interpreting}. As EE-MRIO describes the environmental footprint for global economic activities, it has become the fundamental framework for EE research, illustrated in Figure \ref{fig:EEMRIO}. EE-MRIO supports various studies such as Structure Decomposition Analysis (SDA) and Index Decomposition Analysis (IDA) to identify changes by decomposing key drivers \cite{hoekstra2003comparing, peters2017key, duan2019economic}; monitoring embodied emissions with resource transfer and sustainability evaluation of supply chain in global trade \cite{meng2023narrowing, tian2022regional, wu2020carbon, he2019flow, li2020carbon, long2018embodied, sun2020exploring}. Recently, ML algorithms have been applied with EE-MRIO for several applications, such as accurately identifying ecological hotspots and inefficiencies within the global supply chain to optimise logistic paths for decreased carbon emissions while considering cost-effectiveness \cite{akbari2021systematic}. ML algorithms are naturally suitable for learning multi-dimensional patterns and are utilised for better sectoral sustainability assessment considering environmental, economic and social impacts \cite{abdella2020sustainability, nilashi2019measuring}. Additionally, the ML algorithms extract the patterns from the original high-dimensional space to latent feature space, and it is leveraged to extract the critical paths of products' energy, water, and emission footprints from giant trading networks for supporting sustainable investment decisions \cite{ding2022identifying}.

However, unlike the open-source culture of the ML community \cite{zheng2022multi}, most EE studies haven't publicised the code and data. This leads to fragmented, independent, and inconsistent research leveraging ML techniques to explore the sustainability of different sectors and regions. Subsequent research cannot fairly compare the model performance, such as accuracy, robustness and generalisation or reproduce the result based on previous research, leading to the slow EE-ML development \cite{zhu2023carbonmonitor, ballarin2023climbra, nangini2019global}. Furthermore, no public benchmark EE-ML dataset is available to the best of our knowledge. The EE-MRIO data is a high-quality source for constructing ML-ready datasets. However, the existing dataset faces limitations in terms of accessibility and resolution. Specifically, Eora offers high-quality data but is closed-source and requires high-cost purchasing \cite{lenzen2013building}. Global Trade Analysis Project (GTAP) covers only five reference time steps with non-free access \cite{chepeliev2023gtap}. The World Input-Output Database (WIOD) \cite{dietzenbacher2013construction} has been questioned for its low resolution of temporal and spatial scope, whose latest temporal coverage is 2016. Therefore, uniform problem formulations and benchmark datasets need to be urgently developed. In summary, there exists a big gap in the cooperation between Ecological Economics and Machine Learning (EE-ML), which is demonstrated in three aspects: the inaccessibility of the dataset, the challenge of intensive data preprocessing as the domain knowledge being required and the lack of Benchmark ML datasets and models.

To fill these gaps, we proposed ExioML as the first ML-ready benchmark data in EE research. The ExioML is developed on top of the high-quality open-source EE-MRIO dataset ExioBase 3.8.2 with high spatiotemporal resolution \cite{stadler2018exiobase}. Enabling tracking resource flow between international sectors for global sectoral sustainability assessment. The ExioML developing toolkit is open-sourced, provides Footprint Networks calculation with GPU acceleration, and contributes to the high flexibility of customising interested factors, significantly reducing the barrier for ML research. The usability of the ExioML dataset was validated using multimodality regression to quantify the sectoral GHG emission, achieving a low MSE result. This establishes a critical baseline for accuracy and efficiency in future ML research.

% \begin{table}[h]
%     \centering
%     \caption{Comparison between exsiting EE-MRIO datasets.}
%     \begin{tabular}{ c||c|c|c|c}
%     \hline
%     \textbf{Characteristics} & \textbf{GTAP} & \textbf{OpenEU} & \textbf{WIOD} &  \textbf{ExioML}\\
%     \hline
%     Fully open access & \xmark & \cmark & \cmark & \cmark\\
%     ML adaptation & \xmark & \xmark & \xmark & \cmark \\
%     Geographical coverage & \textbf{141} & 113 & 43 & 49\\
%     Temporal coverage & 5 & - & 15 & \textbf{28} \\
%     Sector coverage & 65 & 128 & 56 & \textbf{363}\\
%     Environmental indicators & - & - & -  & \textbf{1082}\\
%     \hline
%     \end{tabular}
%     \label{tab:data}
% \end{table}

\section{Methods}\label{sec2}
\begin{figure} [h!]
    \centering
    \includegraphics[width=1\linewidth]{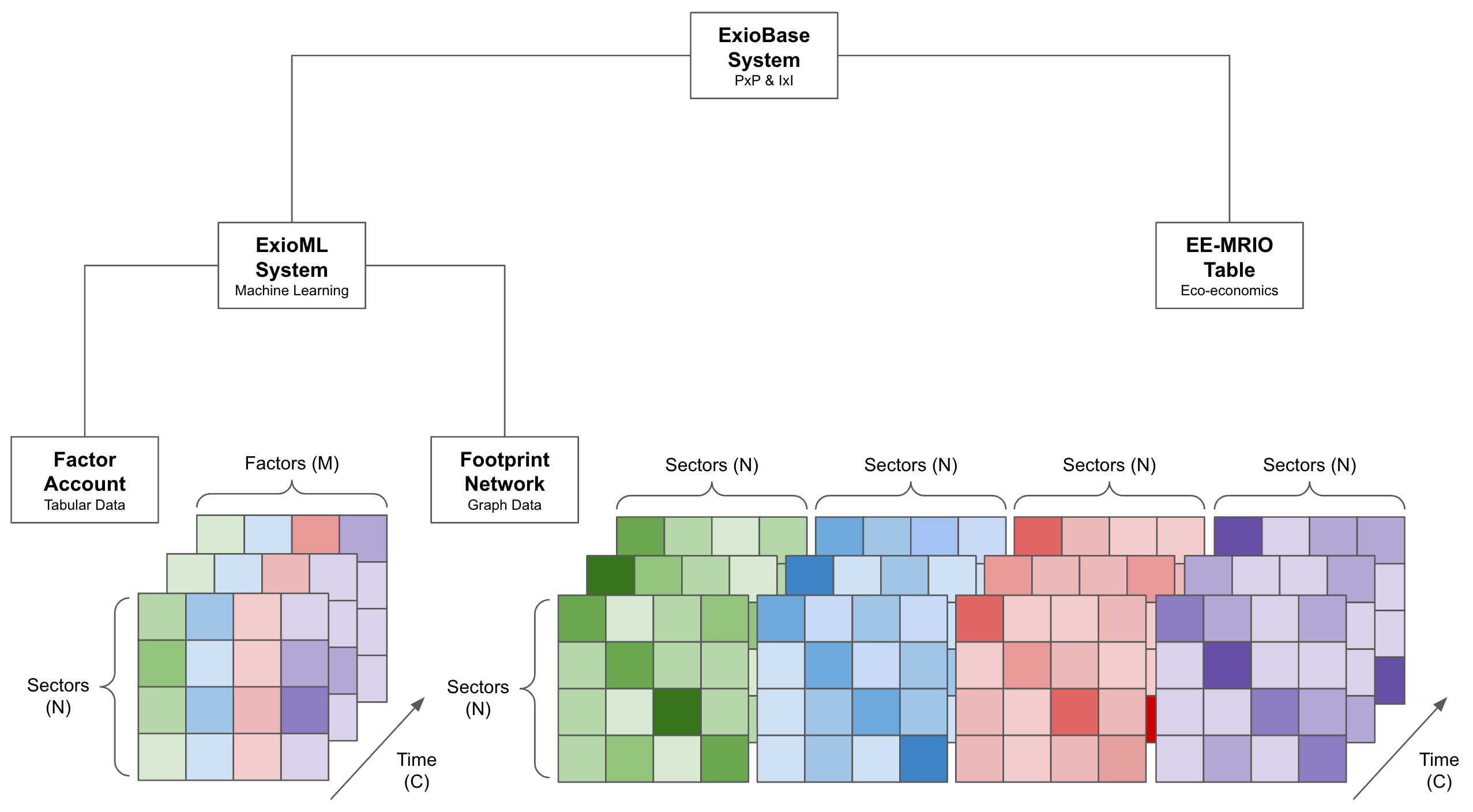}
    \caption{Architecture of ExioML system derived from the open-source EE-MRIO database, ExioBase 3.8.2. Each colour indicates an Eco-economics factor: value added, employment, energy consumption and GHG emission. The system contains Factor Accounting data describing heterogeneous sector features. The Footprint Networks model the global trading network tracking resource transfer within sectors. The data is presented into 2 categories: 200 products and 163 industries for 49 regions from 1995 to 2022 in the PxP and IxI datasets.}
    \label{fig:exioML}
\end{figure}

The architecture of the proposed ExioML dataset is illustrated in Figure \ref{fig:exioML}. It ends the challenges of data limitation and simplifies the EE-MRIO framework for ML researchers in a ready-to-use manner. ExioML offers high spatiotemporal resolution for 49 regions from 1995 to 2022 containing tabular and graphical formats, supporting diverse ML research, such as time-series emissions forecasting \cite{Deb2017review}, factor analysis \cite{raihan2022relationship, zhang2019energy, duan2019economic, acheampong2019modelling, khan2020heterogeneity}, graph learning \cite{sun2020analyzing} and clustering analysis \cite{he2022factors, kijewska2016research}, under a uniform framework. The details of ExioML are summarised in Tablue \ref{tab:Exio}.

\begin{table}[h]
    \centering
    \caption{Summaries of details of ExioML dataset.}
    \begin{tabular}{ c||c }
    \hline
    \textbf{Characteristics} & \textbf{Description} \\
    \hline
    Components & Factor accounting (Tabular data), Footprint Networks (Graph data) \\
    Time frame & Covers 28 annual time steps from 1995 to 2022 \\
    Geographical coverage & 49 regions (44 countries and 5 rest of the world) \\
    Sectors detail & 200 products in PxP, 163 industries in IxI \\
    Key factors & Value added, employment, energy consumption, and GHG emission \\
    \hline
    \end{tabular}
    \label{tab:Exio}
\end{table}

ExioML is derived from an open-source EE-MRIO dataset, ExioBase 3.8.2 \cite{stadler2018exiobase, stadler2021pymrio}, a comprehensive source database covering 417 emission categories and 662 material and resource categories occupying over 40 GB of storage. ExioBase 3.8.2 is available in two versions based on different sector classifications. Specifically, the first version is defined on a product-by-product (PxP) basis, covering 200 products, while the second version is defined on an industry-by-industry (IxI) basis, encompassing 163 industries. ExioML treats products and industries as different sector objects, describes them using common features, and creates PxP and IxI into two complementary datasets. Therefore, ExioML preserves the comprehensive information of raw data sources. ExioML provides PxP and IxI data under a similar distribution for self-contained validation. It enables future ML research to evaluate the generalisation and robustness of ML models on two datasets and reduces the effort to seek datasets under the same structures.

Additionally, we open-sourced the well-documented ExioML development toolkit to stimulate interdisciplinary corporations in three contributions. Firstly, by providing the footprint calculation, the toolkit reduces the barrier for new ML researchers without EE domain knowledge. Secondly, data storage and computational demands are substantial due to high-dimensional matrix operations. ExioML development toolkit utilises GPU accelerations and novelty stores multi-dimensional networks in a single edge table, significantly reducing the computing and storage. Thirdly, the toolkit provides a highly flexible interface for querying the factors of interest from ExioBase 3.8.2. MRIO features contain specific Eco-economics meanings, such as various emissions types. Independent features should be carefully selected. Therefore, four essential factors are identified by the EE research community: GHG emissions, population, gross domestic product (GDP), and total primary energy supply (TPES). The decomposition uses the Kaya Identity, a popular structural decomposition method designed specifically for climate change \cite{peters2017key}. This method breaks down carbon emissions into four indicators, which are population ($P$), GDP per capita ($G/P$), energy intensity ($E/G$) and carbon intensity ($F/E$):

\begin{equation}
    F = P \times \frac{G}{P} \times \frac{E}{G} \times \frac{F}{E}.
\end{equation}

Intensive data preprocessing, such as heavy data cleaning, structure transformation, and Footprint Networks calculation, is required to transfer the raw data into a standardized ML paradigm. Here are the construction details of the ExioML database that utilises the MRIO framework. $Z$ is the transaction matrix that indicates the global inter-industries flows within and across $m$ regions. Each submatrix $Z^{rs}$ represents the trade from the inter-regional requirement from region $r$ to region $s$ for each sector. Similarly, $Y$ is the demand matrix similar. 
Initially, the column vector $x$ symbolizes the global economy's output, with each element $x^r$ representing the total output of region $r$. This could be determined by transaction matrix $Z$, demand matrix $Y$ and $e$ is the summation vector:

\begin{equation}
    \begin{pmatrix}
    x^1 \\
    x^2 \\
    \vdots \\
    x^m \\
    \end{pmatrix} = \begin{pmatrix}
    Z^{11} & Z^{12} & \cdots & Z^{1m} \\
    Z^{21} & Z^{22} & \cdots & Z^{2m} \\
    \vdots  & \vdots  & \ddots & \vdots  \\
    Z^{m1} & Z^{m2} & \cdots & Z^{mm}
    \end{pmatrix} e + \begin{pmatrix}
    Y^{11} & Y^{12} & \cdots & Y^{1m} \\
    Y^{21} & Y^{22} & \cdots & Y^{2m} \\
    \vdots  & \vdots  & \ddots & \vdots  \\
    Y^{m1} & Y^{m2} & \cdots & Y^{mm}
    \end{pmatrix} e.
\end{equation}

Then, the direct requirement matrix $A$, indicative of technological efficiency, is derived by multiplying the transaction matrix $Z$ with $\hat{x}^{-1}$, which is the diagonalized and inverted vector of the global economy's output $x$:

\begin{equation}
    A = Z \hat{x}^{-1}.
\end{equation}

Further, the economy's output vector $X$ is expressible via the Leontief Matrix $L$:

\begin{equation}
    X = (I - A)^{-1} y = Ly.
\end{equation}

This framework can incorporate environmental accounting, such as energy consumption and greenhouse gas (GHG) emissions, with the factor represented by $F$. The coefficient $S$ normalizes this factor against the output $x$:

\begin{equation}
    S = F \hat{x}^{-1}.
\end{equation}

Finally, the Footprint Networks $D$ could be determined by:

\begin{equation}
    D = SLy.
\end{equation}

\section{Data Records}
ExioML supports graph and tabular structure learning algorithms using Footprint Networks and Factor Accounting table. The factors included in PxP and IxI in ExioML are detailed in Table \ref{tab:feature}.

\begin{table}[h]
    \centering
    \caption{Feature of PxP and IxI dataset in ExioML.}
    \begin{tabular}{ c||c|c }
    \hline
    \textbf{Attribute name} & \textbf{Type} & \textbf{Description} \\
    \hline
    Region & Categorical & Region code (e.g. AU, US, CN) \\
    Sector & Categorical & Product or industry (e.g. biogasoline, construction) \\
    Year & Numerical & Timestep (e.g. 1995, 2022)\\
    Value added & Numerical &  GDP expressed in millions of Euros\\
    Employment & Numerical & Population engaged in thousands of persons\\
    Energy carrier net total & Numerical & Sum of all energy carriers in Terajoules\\
    GHG emission & Numerical & GHG emissions in kilograms of CO$_2$ equivalent \\
    \hline
    \end{tabular}
    \label{tab:feature}
\end{table}

\begin{figure} [h!]
    \centering
    \includegraphics[width=0.88\linewidth]{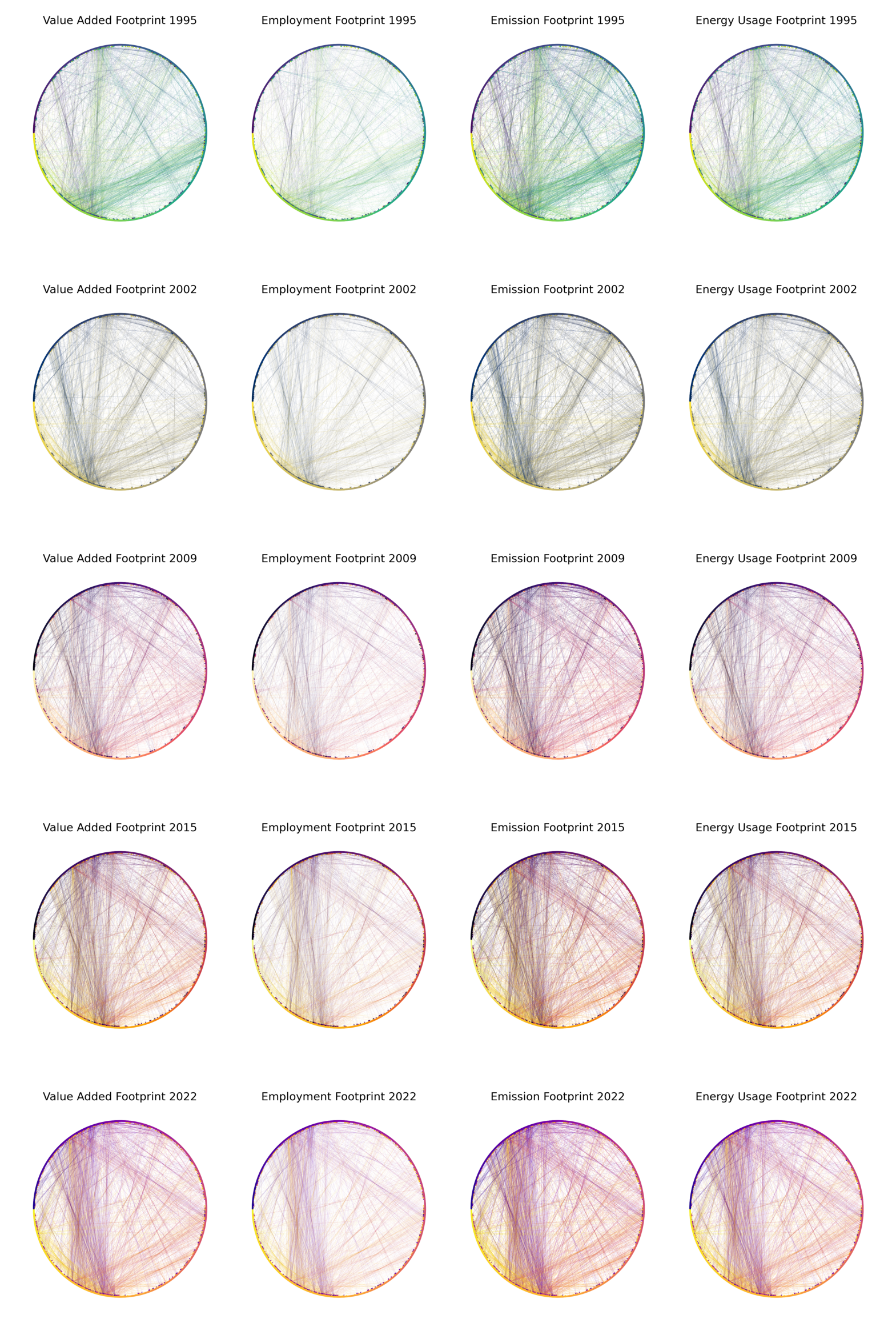}
    \caption{Dynamic Global Trade Footprint Networks from 1995 to 2022. This visualization employs a circular layout to depict the evolving trade patterns over time. Nodes represent sectors indicated by colours. Edges indicate resource transfers, with colours reflecting the source regions. The diagram highlights significant shifts in primary trade sources and target sectors, illustrating the dynamic structural changes in the global trading network.}
    \label{fig:footprint}
\end{figure}

The Footprint Networks model the high-dimensional global trading networks, capturing its economic, social, and environmental impacts illustrated in Figure \ref{fig:footprint}. This network is structured as a directed graph, where the directionality represents the sectoral input-output relationships, delineating sectors by their roles as sources (exporting) and targets (importing). The basic element in the ExioML Footprint Networks is international trade across different sectors with different features such as value-added, emission amount, and energy input. The Footprint Networks' potential pathway impact is learning the dependency of sectors in the global supply chain to identify critical sectors and paths for sustainability management and optimisation. 

\begin{figure} [h!]
    \centering
    \includegraphics[width=0.98\linewidth]{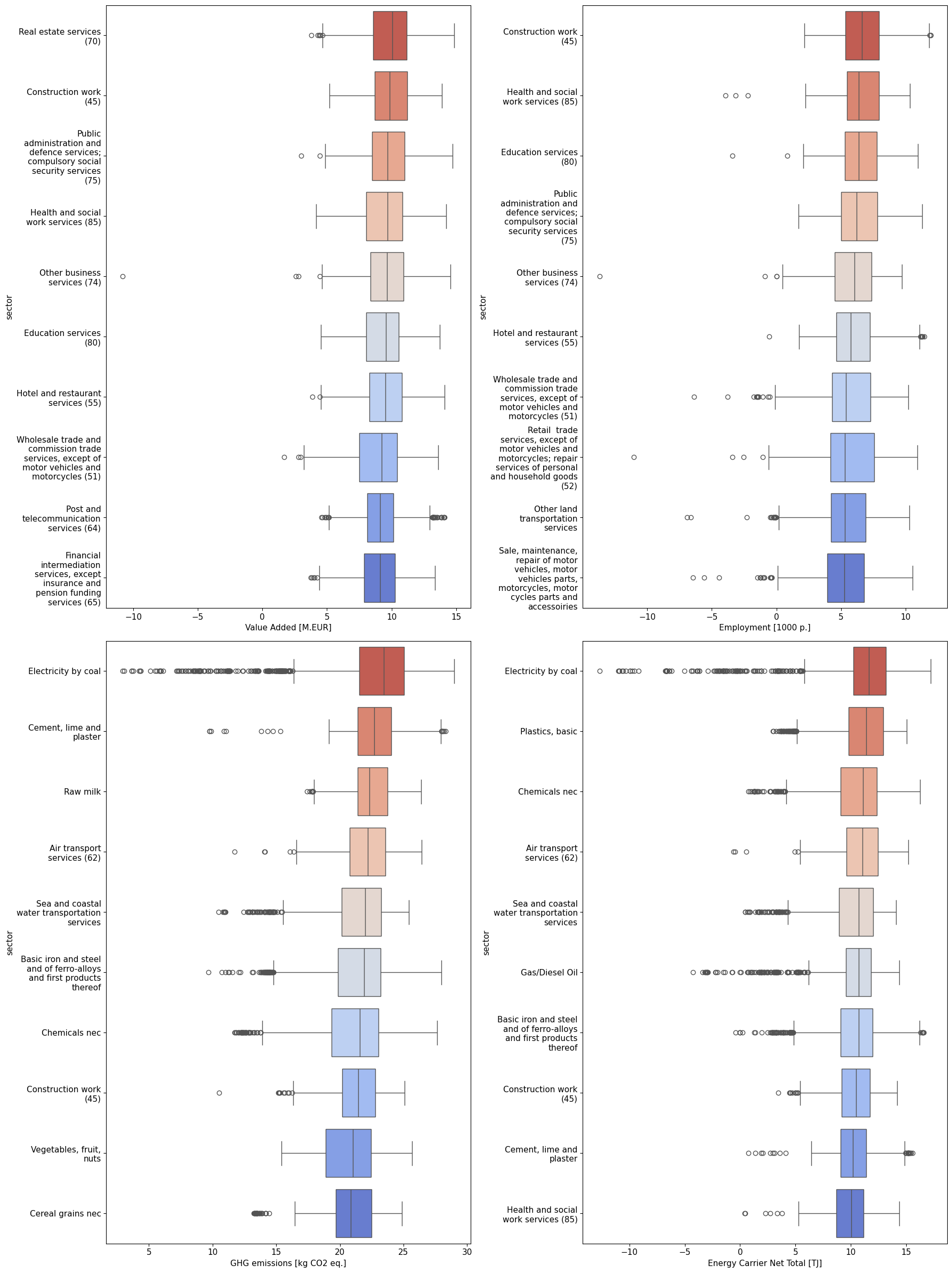}
    \caption{Boxplots for top 10 sectors with largest value-added, employment, GHG emission and energy usage in PxP Factor Accounting table. The x-axis represents indicators transformed using a logarithmic scale, while the y-axis lists sector names. These boxplots reveal the high skewness in sector distributions across different regions.}
    \label{fig:boxplot}
\end{figure}

\begin{figure} [h!]
    \centering
    \includegraphics[width=0.98\linewidth]{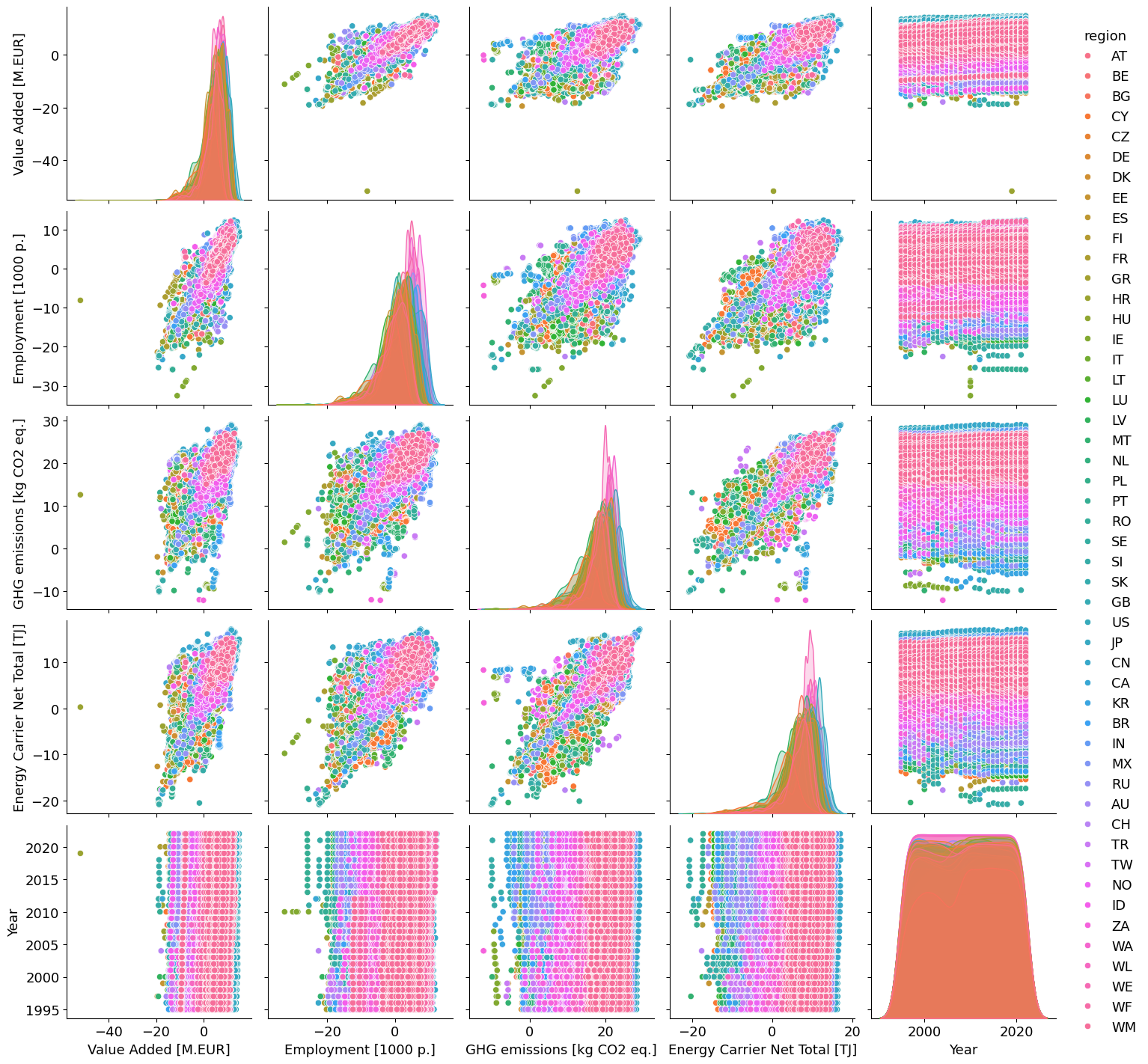}
    \caption{The diagonal subfigures in the pair plot display the sectoral distribution with respect to four selected factors. Off-diagonal subfigures present the pairwise scatter plots, where each point represents a sector and is color-coded by region. The plot reveals high relative pairwise correlations among these factors.}
    \label{fig:pairplot}
\end{figure}

The second part of ExioML is the Factor Accounting table, which shares the common features with the Footprint Networks and summarises the total heterogeneous characteristics of various sectors, which could be leveraged to perform sectoral sustainability assessment. Figure \ref{fig:boxplot} describes the boxplot of the top 10 sectors respective to four selected factors, and Figure \ref{fig:pairplot} displays the pairplot of sector scatters. Log transformation is applied to re-scale the Eco-economics data due to inherent skewness.

Storing large multi-dimensional Footprint Networks poses a significant challenge. For example, a single factor's connections for one year amount to 63,792,169, consuming approximately 1 GB of storage. We addressed this challenge by leveraging the characteristics of the global trading network. By representing the network in an edge table, we stored the Multi-dimensional Footprint Networks in a single file, significantly reducing computing and storage requirements. Given that the global trading network is scale-free and edge weights follow a power-law distribution \cite{foster2005simplistic}, the network is overall sparse, with connections concentrated around dominant nodes. We transformed the graph adjacency matrix into an edge table in tabular format, selecting the top 1,000,000 connections per year. After performing an inner join with the selected factors, around 300,000 edges per year remained. The total storage consumption for the PxP and IxI Footprint Networks is 1.43 GB and 1.34 GB, respectively. The Factor Accounting table requires even less computation and storage, consuming approximately 20 MB.

\section{Technical Validation}
GHG emissions are a crucial socio-economic measure for evaluating sectoral sustainability, yet they remain highly uncertain in terms of quantification \cite{sun2022predictions, riahi2007scenarios}. Emissions are an indirect indicator, unlike value-added, employment, or energy consumption, which can be directly measured. The calculation of sectoral emissions depends on the efficiency of raw materials and production systems, which vary widely across different sectors. Consequently, sectors struggle to disclose their carbon emissions accurately, impacting carbon transparency \cite{matisoff2013different}. Therefore, our study demonstrates a multimodel sectoral GHG emissions regression task leveraging categorical and numerical features to validate the usability of the proposed ExioML. The model aims to automatically learn the underlying relationships of Eco-economics factors to quantify sectoral GHG emissions. 

% The inherent skewness in Eco-economics data presents a significant challenge for analysis. To address this, employing log-transformation or scaling methods is a critical step in data preprocessing for effective machine learning modelling. Figure \ref{fig:distribution} illustrates the distribution of numerical features, while

% Figure \ref{fig:correlation} depicts the correlation coefficients between variables. An analysis of these figures reveals a strong positive Pearson correlation among value added, employment, energy, and emissions. In contrast, feature year demonstrates a lower positive correlation with the other features. 

% \begin{figure} [h]
%     \centering
%     \includegraphics[width=1\linewidth]{figure/distribution.png}
%     \caption{Distribution of Factor Accounting tabe in PxP and IxI.}
%     \label{fig:distribution}
% \end{figure}

% \begin{figure} [h]
%     \centering
%     \includegraphics[width=1\linewidth]{figure/correlation.png}
%     \caption{Pearson correlation matrix of Factor Accounting tabe in PxP and IxI.}
%     \label{fig:correlation}
% \end{figure}

The supervised regression task is to learn the mapping function $f_{\theta}: X \mapsto Y$ from feature vector $x_i \in X$ to labels $y_i \in Y$ on dataset $\mathcal{D} = \{(x_1, y_1), ..., (x_n, y_n)\}$. In this case, we utilized the PxP and IxI as dataset $\mathcal{D}$ of ExioML's Factor Accounting table, comprising 221,307 and 179,185 instances after excluding missing values, respectively. The feature vector $X$ is value-added, employment and energy consumption, while GHG emission is set as the label $Y$. 

Regression models could be categorized into shallow and deep learning models. Deep learning in Eco-economics analysis, particularly on tabular data, remains relatively underexplored \cite{gorishniy2021revisiting}. We rigorously evaluated different learning algorithms on tabular learning algorithms, including tree-based models, MLP and Transformer architectures \cite{vaswani2017attention}. The distinction lies in their handling of categorical features: shallow models employ Leave-One-Out Text Encoding due to their inability to process categorical data natively \cite{zhang2016tips}, whereas deep learning models learn categorical embeddings directly from raw data. We implemented the deep learning algorithms by PyTorch Tabular package \cite{joseph2021pytorch}. Algorithms included in GHG emission regression are as follows:

\begin{itemize}
    \item \textbf{K-Nearnest-Neighbour (KNN):} This non-parametric method predicts based on the majority vote from its k nearest neighbours \cite{cover1967nearest}.
    \item \textbf{Ridge Regression (Ridge):} A linear regression variant employing $\ell_2$ regularization to prevent overfitting \cite{hoerl1970ridge}.
    \item \textbf{Decision Tree (DT):} A feature splitting algorithm to tree flowchart structure for decision making \cite{quinlan1986induction}.
    \item \textbf{Random Forest (RF):} Ensemble of multiple decision trees for robust prediction \cite{breiman2001random}.
    \item \textbf{Grandient Boost Decision Tree (DBDT):} Ensemble weak trees to improve the error of the previous tree sequentially \cite{friedman2001greedy}. 
    \item \textbf{Multi-Layer Perceptron (MLP):} A deep artificial neural network employing backpropagation with the capability of non-linear function approximation \cite{lecun2015deep}.
    \item \textbf{Gated Adaptive Network (GANDALF):} GANDALF leverages Gated Feature Learning Units (GFLU) for automated feature selection \cite{joseph2022gate}.
    \item \textbf{Feature-Tokenizer Transformer (FTT):} Novel transformer architecture applies a tokenization mechanism on tabular features \cite{gorishniy2021revisiting}.
\end{itemize}

This analysis assessed each model based on its mean square error and training time. To ensure unbiased evaluation and prevent leakage of target information, the dataset was partitioned into training, validation, and testing sets with proportions of 64\%, 16\% and 20\% to avoid label information leakage. Optimal model parameters were determined through a random search across 30 trials on the training and validation data. We conducted 10 experiments for each tuned model hyperparameter with their performance on the test set. The experimental procedures are outlined in Table \ref{tab:details}. Data normalization was employed for DL models to ensure stable training, although this did not significantly enhance performance. In terms of implementation, the MPS framework was used to train the deep models on Apple M3 Pro GPUs. 
\begin{table}[h]
    \centering
    \caption{Impletement detail.}
    \begin{tabular}{ c|c||c|c }
    \hline
    \textbf{Data} & \textbf{Description} & \textbf{Setting} & \textbf{Description} \\
    \hline
    \# Train & 141,635 (PxP), 114,678 (IxI) & Epoch & 30\\
    \# Validation & 35,409 (PxP), 28,670 (IxI) & Batch size & 512 \\
    \# Test & 44,262 (PxP), 35,837 (IxI) & Shallow & Sci-kit Learn\\
    Encoding & Shallow models only & Deep & PyTorch Tabular\\
    Normalisation & DL models only& Version & Python 3.11.7 \\
    Scaling & Log-transformation & GPU & Apple M3 Pro 18 GB \\
    \hline
    \end{tabular}
    \label{tab:details}
\end{table}

A random search of hyperparameters was conducted over 30 trials within a defined parameter grid. The range of parameters tested and the best parameters identified are summarized in Table \ref{tab:param}. The search grid was heuristically small and based on model characteristics and previous studies to optimize efficiency. The tuning process for each ensemble model took about 10 to 20 minutes per dataset, whereas for each DL model, it took roughly 2 to 3 hours.

\begin{table}[h]
    \centering
    \caption{Parameter grid for tuning and best parameter found for PxP and IxI.}
    \begin{tabular}{ c||c|c|c }
    \hline
    \textbf{Model} & \textbf{Parameter range} & \textbf{Best PxP} & \textbf{Best IxI} \\
    \hline
    KNN & & \\
    \# Neighbours & Range(1, 50, 2) & 5 & 3\\
    Weights & Uniform, Distance & Distance & Distance\\
    Metric & Euclidean, Manhattan & Manhattan & Manhattan\\
    \hline
    Ridge & &\\
    Alpha & 0, 0.001, 0.01, 0.1, 0, 1, 10, 100 & 0 & 100\\
    \hline
    DT & &\\
    Max depth & Range(1, 50, 2) & 46 & 46\\
    Min samples leaf & 1, 2, 3 & 3 & 1\\ 
    Min samples split & 2, 4, 6, 8 & 4 & 6 \\
    Max features & Sqrt, Log2 & Sqrt & Log2\\
    \hline
    RF & &\\
    Max depth & Range(1, 50, 2) & 26 & 26\\
    Min samples leaf & 1, 2, 3 & 1 & 1\\ 
    Min samples split & 2, 4, 6, 8 & 2& 2\\
    Max features & Sqrt, Log2 & Sqrt & Log2 \\
     \# Estimators & 50, 100, 150 & 100 & 100\\
    \hline
    GBDT & &\\
    Max depth & Range(1, 50, 2) & 16 & 31\\
    Min samples leaf & 1, 2, 3 & 3 & 3 \\ 
    Min samples split & 2, 4, 6, 8 & 8 & 2\\
    Max features & Sqrt, Log2 & Log2 & Log2 \\
    \# Estimators & 50, 100, 150 & 100 & 100 \\
    Learning rate & 0.01, 0.1, 1 & 0.1 & 0.1\\
    \hline
    MLP & &\\
    Layers & 256-128-64, 128-64-32, 64-32-16 & 256-128-64 & 256-128-64 \\
    Dropout & 0, 0.05, 0.1 & 0 & 0 \\
    Learning rate & 0.001, 0.01 & 0.01 & 0.01\\
    \hline
    GANDALF & & \\
    GFLU stages & 1,2,3,4,5 & 4 & 4 \\
    GFLU dropout & 0, 0.05, 0.1 & 0 & 0 \\
    Feature sparsity & 0, 0.05, 0.1 & 0 & 0 \\
    Learning rate &  0.001, 0.01 & 0.01 & 0.01 \\
    \hline
    FTT & & \\
    \# Heads & 4, 8 & 8 & 8 \\
    \# Attention blocks & 1, 2, 3, 4 & 2 & 3 \\
    \# Multiplier & 1, 2, 3, 4 & 4 & 1 \\
    Learning rate & 0.001, 0.01 & 0.01 & 0.01\\
    \hline
    \end{tabular}
    \label{tab:param}
\end{table}

Table \ref{tab:result} summarized the model's mean performance and standard deviation. The primary finding is that deep learning models exhibit marginally lower mean squared errors than ensemble counterparts. GANDALF shows the most effective performance in PxP and IxI data. Ensemble methods, GBDT and RF, emerge with competitive accuracy and less demand for computational resources. Although deep models benefit from GPU acceleration, their training time is significantly longer than shallow models. We demonstrated the evaluation of various machine learning models on Factor Accounting datasets focused on sectoral GHG emission prediction through regression tasks, serving as effective baselines with low mean square error performance. 

\begin{table}[!h]
    \centering
    \caption{Result for models with standard deviation for 10 runs and the top results are in \textbf{bold}.}
    \begin{tabular}{ c||c|c||c|c }
    \hline
    & \multicolumn{2}{c||}{\textbf{PxP}} & \multicolumn{2}{c}{\textbf{IxI}}\\
    \hline
    Model & MSE & Time (s)  & MSE & Time (s) \\
    \hline
    KNN & 1.071 ± 0.010 & 0.035 ± 0.001 & 1.151 ± 0.018 & 0.026 ± 0.001\\
    Ridge & 2.265 ± 0.000 & \textbf{0.005 ± 0.002} & 2.514 ± 0.000 & \textbf{0.004 ± 0.002}\\
    DT & 0.926 ± 0.051 & 0.316 ± 0.017 & 0.848 ± 0.070 & 0.254 ± 0.019\\
    RF & 0.356 ± 0.004 & 21.521 ± 0.157 & 0.302 ± 0.004 & 16.511 ± 0.060\\
    GBDT & 0.234 ± 0.006 & 30.276 ± 0.224 & 0.219 ± 0.007 & 32.847 ± 0.388\\
    MLP & 0.226 ± 0.007 & 219.218 ± 0.904 & 0.250 ± 0.092 & 205.051 ± 24.309\\
    GANDALF & \textbf{0.204 ± 0.010} & 352.756 ± 7.036 & \textbf{0.189 ± 0.007} & 383.119 ± 3.664\\
    FTT & 0.330 ± 0.007 & 330.578 ± 1.527 & 0.302 ± 0.023 & 468.911 ± 7.329\\
    \hline
    \end{tabular}
    \label{tab:result}
\end{table}

To summarize, this research introduces the first high-quality EE-ML benchmark dataset integrated with the EE-MRIO framework, a powerful analysis tool in ecological economics research. The novel ExioML dataset and development toolkit address three significant challenges in sustainability research and establish benchmark data and algorithms for future studies. Firstly, ExioML overcomes data access limitations by providing high-resolution resources. Secondly, it streamlines the complexity of the EE-MRIO framework with GPU-accelerated computations and efficient storage management, offering high-level factor customization for future researchers. Finally, we demonstrated a sectoral GHG emission prediction task to the usability of the proposed dataset on various machine learning models and achieved low mean square errors. 

ExioML fosters interdisciplinary collaboration between ecological economics and machine learning, laying a solid foundation for further research in areas such as supply chain optimization, sustainability assessment, and footprint prediction. This contributes to formulating effective climate policies and making sustainable investment decisions.

\section{Code Availability}
The ExioML development toolkit by Python and the regression model used for validation are available on the GitHub repository (\url{https://github.com/Yvnminc/ExioML}). And the ExioML data is hosted by Zenodo (\url{https://zenodo.org/records/10604610}).




% \end{appendices}


\bibliography{sn-bibliography}% common bib file

\clearpage
\appendix
\section{Competing interests}
The authors declare no competing interests.

\section{Author Contributions}
Y. Guo designed the study and produced the dataset, visualisations, and regression models for technical validation. J. Ma supervised the project. Both authors participated in writing the paper. C. Guan participated in the project design discussion and helped improve the paper draft.

\end{document}